\documentclass[a4paper,11pt,fleqn]{cas-sc}
\renewcommand{\printorcid}{}

\usepackage[authoryear]{natbib}
\usepackage{setspace}
\usepackage{microtype}
\usepackage{graphicx}
\usepackage{subcaption}
\usepackage{caption}
\usepackage{booktabs} 
\usepackage{hyperref}
\usepackage[most]{tcolorbox} 
\usepackage{xcolor} 
\usepackage{multicol}
\usepackage{multirow}
\usepackage{listings}
\usepackage{stfloats}
\usepackage{textcomp}
\newtcolorbox{ORAgentBox}[1]{ enhanced,breakable, colback=blue!5, colframe=gray!80, fonttitle=\bfseries, title=#1, drop shadow, attach boxed title to top left={xshift=3mm, yshift=-2mm}, boxed title style={colback=gray!80}, left=3mm, right=3mm, top=1mm, bottom=1mm }
\newtcolorbox{mygraybox}[1]{ enhanced, title=#1, fonttitle=\bfseries, fontupper=\small, colback=blue!5, colframe=gray!50, coltitle=black, attach boxed title to top left={yshift=-2mm, xshift=5mm}, boxed title style={colback=white, colframe=gray!50, boxrule=0.5pt}, arc=2mm, boxrule=1pt, left=0.5mm, right=0.5mm, top=1mm, bottom=0.5mm, before skip=10pt, after skip=10pt }
\usepackage{varwidth}

\def\tsc#1{\csdef{#1}{\textsc{\lowercase{#1}}\xspace}}
\tsc{WGM}
\tsc{QE}

\begin{document}
\let\WriteBookmarks\relax
\def\floatpagepagefraction{1}
\def\textpagefraction{.001}

\shorttitle{}    

\shortauthors{Y. Shi et~al.}  

\title [mode = title]{MIRROR: A Multi-Agent Framework with Iterative Adaptive Revision and Hierarchical Retrieval for Optimization Modeling in Operations Research}  

\affiliation[1]{organization={School of Mathematics and Statistics, Xi'an Jiaotong University},
	addressline={No.28, Xianning West Road},
	city={Xi'an},
	state={Shaanxi},
	postcode={710049},
	country={China}}

\affiliation[2]{organization={School of Electronics and Information, Northwestern Polytechnical University},
addressline={No.1, Dongxiang Road},
	city={Xi'an},
	state={Shaanxi},
	postcode={710129},
	country={China}}

\author[1]{Yifan Shi}%
\fnmark[1]
\ead{syf123456@stu.xjtu.edu.cn}
\credit{Conceptualization; Data curation; Investigation; Methodology; Software; Writing – original draft; Writing – review \& editing; Validation; Visualization}

\author[1]{Jiayi Wang}%
\fnmark[1]
\ead{wangjiayi21@stu.xjtu.edu.cn}
\credit{Writing – original draft; Visualization; Investigation; Methodology; Software; Validation; Writing – review \& editing; Data curation}

\author[1]{Minyi Wu}%
\ead{3125307075@stu.xjtu.edu.cn}
\credit{Data curation; Investigation; Software; Validation}
\author[2]{Ye Fan}%
\ead{fanye@nwpu.edu.cn}
\credit{Data curation; Investigation; Software; Validation}
\author[1]{Jialong Shi}%
\cormark[1]
\ead{jialong.shi@xjtu.edu.cn}
\credit{Project administration; Supervision; Writing – review \& editing; Resources}
\author[1]{Jianyong Sun}%
\cormark[1]
\ead{jy.sun@xjtu.edu.cn}
\credit{Project administration; Supervision; Writing – review \& editing; Resources}
\fntext[1]{Equal contribution.}

\cortext[1]{Corresponding authors.}

\begin{abstract}
Operations Research (OR) relies on expert-driven modeling—a slow and fragile process ill-suited to novel scenarios. While large language models (LLMs) can automatically translate natural language into optimization models, existing approaches either rely on costly post-training or employ multi-agent frameworks, yet most still lack reliable collaborative error correction and task-specific retrieval, often leading to incorrect outputs. We propose MIRROR (a Multi-agent framework with Iterative adaptive Revision and hierarchical Retrieval for optimization modeling in Operations Research), a fine-tuning-free, end-to-end multi-agent framework that directly translates natural language optimization problems into mathematical models and solver code.
MIRROR integrates two core mechanisms: (1) execution-driven iterative adaptive revision for automatic error correction, and (2) hierarchical retrieval to fetch relevant modeling and coding exemplars from a carefully curated exemplar library. Experiments show that MIRROR outperforms existing methods on standard OR benchmarks, with notable results on complex industrial datasets such as ``IndustryOR'' and ``Mamo-ComplexLP''. By combining precise external knowledge infusion with systematic error correction, MIRROR provides non-expert users with an efficient and reliable OR modeling solution, overcoming the fundamental limitations of general-purpose LLMs in expert optimization tasks.
\end{abstract}


\maketitle

\section{Introduction}
Operations Research (OR) serves as a foundational methodology for solving complex decision-making problems and plays an indispensable role in domains such as manufacturing, logistics, supply chain management, energy scheduling, and public services~\citep{cannas2024artificial}. By formulating mathematical optimization models and leveraging high-performance solvers such as Gurobi\citep{gurobi2024}, COPT\citep{ge2024cardinaloptimizercoptuser}. OR significantly improves resource utilization efficiency, reduces operational costs, and enables enterprises to maximize economic returns. Despite its well-established value, the practical adoption of OR faces a fundamental bottleneck: real-world problems are typically expressed in unstructured natural language, while translating them into rigorous mathematical models and executable code requires deep domain expertise and programming proficiency, making the process time-consuming, labor-intensive, and poorly scalable. This high knowledge barrier limits OR’s accessibility for small- and medium-sized enterprises and non-technical users, and further complicates adaptation to dynamic changes in operational environments—such as demand shifts or updated constraints—that necessitate frequent, expert-driven model revisions.

Recent advances in large language models (LLMs) have demonstrated remarkable capabilities in natural language understanding, mathematical reasoning, symbolic manipulation, and code generation—often approaching or even surpassing human expert performance. Current mainstream LLMs can be broadly categorized into two paradigms: general models and reasoning models. General models, including GPT-4~\citep{openai2024gpt4technicalreport}, DeepSeek-V3~\citep{deepseekai2025deepseekv3}, and Qwen-3~\citep{yang2025qwen3}, excel in semantic comprehension and broad knowledge coverage through extensive pre-training on textual corpora, enabling accurate interpretation of complex problem descriptions and the generation of preliminary modeling insights. Meanwhile, reasoning model architectures such as GPT-o1~\citep{openai2024openaio1card} and DeepSeek-R1~\citep{guo2025deepseek} explicitly enhance systematic thinking, multi-step reasoning chains, and self-correction mechanisms, thereby achieving superior performance in structured tasks such as mathematical derivation, constraint formalization, and code synthesis.

Motivated by recent advances in large language models (LLMs), researchers have explored various approaches to automatically translate unstructured natural language problem statements into solvable mathematical optimization models, aiming to democratize operations research for non-experts. Recent efforts such as LLMOPT~\citep{jiang2024llmopt}, ORLM~\citep{huang2025orlm}, and MiniOpt~\citep{anonymous2026miniopt} train specialized models via supervised fine-tuning or reinforcement learning on synthetic datasets; however, they face two fundamental challenges: high-quality annotated data is scarce and expensive to construct, and the outputs inherently lack natural verifiability, making evaluation and debugging difficult—particularly for smaller-scale models. To circumvent these limitations, a growing line of work has shifted toward multi-agent frameworks that coordinate multiple LLM agents to collaboratively construct optimization models without any additional training. Systems like Chain-of-Experts~\citep{xiao2023chain}, OptiMUS~\citep{ahmaditeshnizi2024optimus}, ORMind~\citep{wang2025ormind}, and OptiTree~\citep{liu2026optitree} decompose complex modeling tasks into specialized roles and enable iterative interaction among agents, offering a flexible and practical alternative. Nevertheless, existing multi-agent frameworks are still constrained by closed architectures that lack extensibility; their external knowledge is often generated by the large models themselves, introducing hallucinations, biases, or misalignments with task requirements, which yields low-quality and poorly relevant contextual support. More critically, most of these frameworks lack reliable correction mechanisms, making it difficult to detect and rectify errors after code execution, thereby creating hidden risks of solution failure and diminished system credibility.

To address these limitations, we propose MIRROR, a \textbf{M}ulti-agent framework with \textbf{I}terative adaptive \textbf{R}evision and hierarchical \textbf{R}etrieval for optimization modeling in \textbf{O}perations \textbf{R}esearch, with the following key contributions. 

\begin{itemize}
	\item \textbf{We propose MIRROR, an end-to-end multi-agent framework.} This framework requires no fine-tuning and automatically transforms natural language descriptions of optimization problems into executable solver code. Its dual-memory architecture consists of local memory and shared global memory: the former records the outputs required by the revision-stage agents to ensure intra-task consistency, while the latter enables cross-task knowledge transfer.
	
	\item \textbf{We design an Iterative Adaptive Revision (IAR) mechanism.} Whenever solver code execution fails, the mathematical modeling and code generation agents switch to their respective revision experts, diagnose errors in either the model or the solver code, and generate structured revision tips. The system then iteratively refines both the model and solver code without human intervention until a correct solution is obtained or a preset limit reached. Unlike existing methods, our approach stores the historical model, code, and associated revision tips in a local memory pool to provide contextual history for subsequent corrective iterations.
	
	\item \textbf{We propose a Hierarchical Retrieval-Augmented Generation (HRAG) mechanism.} It is based on an exemplar library constructed via an automated synthesis and labeling pipeline. The mechanism employs a two-stage retrieval strategy—first coarse-grained filtering by overall problem semantics and metadata, then fine-grained reranking based on subproblem types and deep semantic similarity—to provide highly relevant exemplar contexts for modeling and code generation, significantly improving model rationality and solver code correctness.
\end{itemize}
MIRROR achieves state-of-the-art performance among current multi-agent approaches on multiple operations research benchmarks. It notably outperforms existing methods on challenging datasets such as ``IndustryOR'' and ``Mamo-ComplexLP''. The framework also demonstrates effectiveness with small open-source language models, enhancing their optimization modeling capabilities without any task-specific training.
\section{Related Work}
\paragraph{LLMs for Math and Code Generation}In recent years, large language models (LLMs) have achieved remarkable advances in mathematical reasoning and code generation. On the mathematical side, specialized models such as Qwen2.5-Math~\citep{yang2024qwen25mathtechnicalreportmathematical} and DeepSeek-Prover~\citep{ren2025deepseekproverv2advancingformalmathematical} have enhanced formalized inference, while systems like AlphaEvolve~\citep{novikov2025alphaevolvecodingagentscientific} and MM-Agent~\citep{liu2025mmagent} further integrate LLMs into end-to-end mathematical problem-solving pipelines. In code generation, frameworks including CodeAct~\citep{pmlr-v235-wang24h}, KareCoder~\citep{huang2024knowledgeawarecodegenerationlarge}, SWE-bench~\citep{jimenez2024swebench}, and Web-bench~\citep{xu2025webbenchllmcodebenchmark} demonstrate the potential of LLMs for autonomous programming and debugging. Building on these developments, the present study focuses on the intersection of mathematical modeling and programming—automated optimization modeling—and categorizes existing solutions into two paradigms: Learning-based LLM Optimization Modeling and Agent-based LLM Optimization Modeling.

\paragraph{Learning-based LLM Optimization Modeling}
Recent learning-driven studies enhance LLM-based optimization modeling via data synthesis, targeted fine-tuning, and reinforcement learning to address domain adaptation and data scarcity in operations research.
ORLM~\citep{huang2025orlm}, OptMATH~\citep{lu2025optmath}, Step-Opt~\citep{wu2025step}, and ReSocratic~\citep{yang2025optibench} are recent works on the data synthesis front. On the learning mechanism side, LLMOPT~\citep{jiang2024llmopt}, MiniOpt~\citep{anonymous2026miniopt}, and SIRL~\citep{chen2025solverinformed} reduce modeling hallucinations and improve generalization.  
OR-R1~\citep{ding2025or} achieves similar effects via verifiable learning mechanisms. 
Additionally, CALM~\citep{tang2025calm} and StepORLM~\citep{zhou2025steporlm} refine reasoning trajectories through corrective adaptation and process supervision. 

\paragraph{Agent-based LLM Optimization Modeling} 
To overcome the inherent limitations of fixed decomposition strategies, agent-based frameworks enhance modeling performance through specialized role allocation and collaborative mechanisms. Chain-of-Experts~\citep{xiao2023chain} introduces a Conductor to orchestrate domain experts, whereas OptiMUS~\citep{ahmaditeshnizi2024optimus} utilizes a modular structure to decouple formulation, coding and evaluation modules. At the execution level, OptimAI~\citep{thind2026optimai} incorporates a Planner and a Code Critic to enable strategic reflection, while OR-LLM-Agent~\citep{zhang2025or} leverages the capabilities of reasoning LLMs to decompose the task into three sub-tasks: modeling, code generation and debugging. Furthermore, to enhance reliability, ORMind~\citep{wang2025ormind} draws on cognitive dual-process theory to implement counterfactual reasoning for error detection, while LEAN-LLM-OPT~\citep{liang2026llmlargescaleoptimizationmodel} employs a lightweight few-shot approach to reduce computational overhead without compromising modeling effectiveness. To move beyond predefined steps, OptiTree~\citep{liu2026optitree} introduces a hierarchical thought generation framework that employs tree search to adaptively decompose complex optimization problems into simpler subproblems based on a structured modeling tree. 

\paragraph{Summary and Gaps} Despite recent progress, existing methods face two main limitations. First, learning-based models rely on annotated datasets that are scarce and expensive to create. Their ``black-box'' nature also makes it difficult to diagnose errors or adapt to new constraints without costly re-training. Second, most current multi-agent frameworks often hallucinate due to a lack of external domain knowledge. Moreover, they typically struggle to effectively use solver feedback for error correction, leading to repeated execution failures. To bridge these gaps, we propose MIRROR, which integrates external knowledge retrieval with execution-based iterative revision.

\begin{figure}[pos=htbp]
	\centering
	\includegraphics[width=1\linewidth]{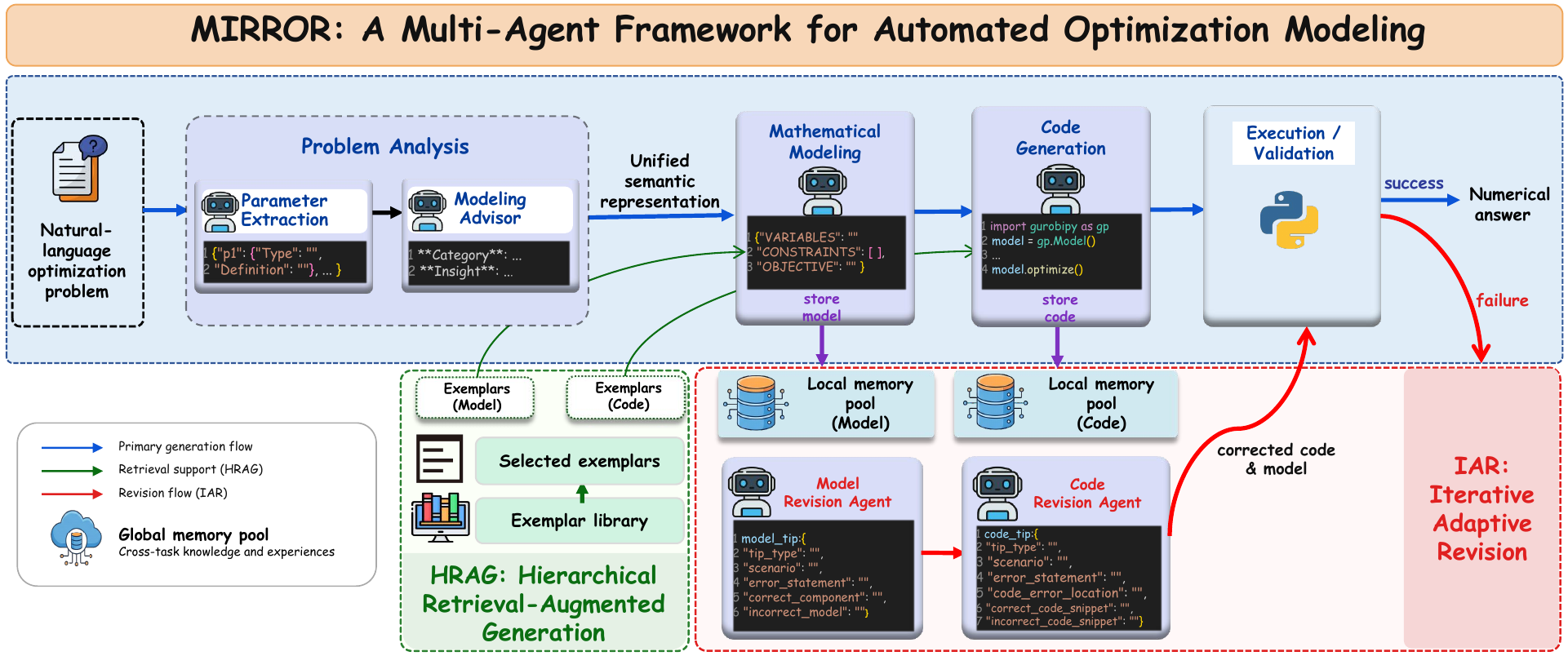}
	\caption{MIRROR: An LLM-based multi-agent framework that automates end-to-end optimization modeling—from natural language to executable solver code—through four phases: Understanding, Modeling, Implementation, and Revision. Hierarchical Retrieval-Augmented Generation (HRAG) retrieves relevant exemplars for model and solver code synthesis; upon execution failure, the Iterative Adaptive Revision (IAR) mechanism leverages local memory to diagnose and refine outputs. Local memory stores per-task agent history for revision, while global memory accumulates cross-task knowledge for system-wide evolution.}
	\label{fig:mirror_pipeline}
	\vspace{-0.4cm}
\end{figure}
\section{Methodology}
\label{eng:methodology}
The core of MIRROR is a synergistic multi-component methodology. MIRROR introduces a carefully curated exemplar library containing problem types and subproblem types to provide detailed exemplar references for the modeling and programming processes. A shared global memory pool stores the outputs produced by all agents throughout the workflow and allows these agents to retrieve previously generated content before producing new outputs. In addition to this shared memory, MIRROR introduces two local memory pools for the modeling and programming processes. These pools exclusively store the corresponding content and revision tips generated during the generation and revision stages, and each local pool can only be accessed and updated by its associated agents. The overall system follows a closed-loop paradigm consisting of analysis, modeling, implementation, and revision to solve optimization problems end to end. The correctness and robustness of the generated results are further supported by Iterative Adaptive Revision (\textsc{IAR}) and Hierarchical Retrieval-Augmented Generation (\textsc{HRAG}). The overall architecture of MIRROR is depicted in Figure \ref{fig:mirror_pipeline}.

To achieve end-to-end automated modeling from natural-language problem descriptions to executable optimization programs, we design a multi-agent framework in which multiple specialized agents collaborate with one another. The complete workflow consists of a generation phase and a revision phase.

\subsection{Generation Phase}
\label{eng:generation}

Given a natural-language optimization problem $t$, the \textbf{Parameter Extraction Agent} ($f_{\text{param}}$), \textbf{Modeling Advisor Agent} ($f_{\text{adv}}$), \textbf{Mathematical Modeling Agent} ($f_{\text{model}}$), and \textbf{Code Generation Agent} ($f_{\text{code}}$) sequentially perform their respective functions during the generation phase, ultimately producing executable solver code $c$.

First, $f_{\text{param}}$ identifies and structures the core parameters of the problem, producing $p$, a JSON object that contains parameter symbols, data types, and semantic definitions. This output is then passed to $f_{\text{adv}}$ as contextual information. The advisor provides domain-aware semantic guidance by generating $g$, a standardized JSON list containing operational explanations of domain-specific terminology, key problem details, and a characterization of the essential structure of the problem.

Subsequently, $f_{\text{model}}$ integrates the original problem $t$, the extracted parameters $p$, the advisory guidance $g$, and the retrieved modeling exemplar set $\mathcal{R}_{\text{model}}$ to construct a formal mathematical model $m$. Finally, $f_{\text{code}}$ translates $m$ into an executable program, handles implementation details, and resolves potential inconsistencies between the abstract formulation and the syntax required by solvers such as Gurobi. Both the Mathematical Modeling Agent and the Code Generation Agent employ the unified \textsc{HRAG} mechanism described in Section~\ref{eng:hrag}. This mechanism dynamically retrieves and incorporates the most relevant modeling paradigms and code exemplars $\mathcal{R}_{\text{code}}$ from the exemplar library.

The external executor receives the generated code $c$ and attempts to run it. It returns either a numerical solution $n$ or a failure flag $\bot$ accompanied by a specific error message $e$, such as a syntax error or an execution timeout. This feedback signal determines whether the revision phase should be triggered. If execution succeeds, the solving process terminates immediately; otherwise, the system enters the iterative revision phase.

\subsubsection{Hierarchical Retrieval-Augmented Generation (\textsc{HRAG})}
\label{eng:hrag}
High-quality exemplar libraries are a critical resource for leveraging in-context learning across diverse domains, as LLM performance is highly sensitive to the demonstrations provided in the prompt~\citep{liu2022good}. Recent studies demonstrate the value of structured knowledge repositories for automated mathematical and optimization modeling. MM-Agent constructs a three-level Hierarchical Mathematical Modeling Library that organizes domains, subdomains, and modeling methods, enabling task-specific method retrieval for subtask formulation~\citep{liu2025mmagent}. Similarly, OptiTree organizes operations research problems by taxonomy and complexity in a modeling tree, stores modeling thoughts at its nodes, and retrieves relevant subproblems and thoughts to guide the decomposition of unseen complex instances~\citep{liu2026optitree}. These designs show that reusable modeling knowledge should be explicitly organized around problem structure, rather than relying solely on the latent knowledge of an LLM. However, the retrieved knowledge in these systems primarily consists of high-level methods or modeling thoughts; it is not designed as a stage-specific library of verified end-to-end examples that jointly support mathematical formulation and solver-code generation. MIRROR addresses this gap by constructing a curated exemplar library specifically designed to support both modeling and code-generation stages.

The \textsc{HRAG} mechanism follows an ``exemplar construction--exemplar allocation'' strategy. During exemplar construction, the system builds an exemplar library $\mathcal{L}$ containing 602 high-quality and distribution-balanced optimization instances. Through context augmentation, each exemplar contains a complete mathematical modeling process and its corresponding solver code, organized as a triplet $(t,m,c)$. Each instance is additionally annotated with a high-level problem category and a fine-grained subproblem type. During exemplar allocation, the system retrieves relevant exemplars and delivers them to the corresponding agents.

\paragraph{\textbf{Exemplar Library Construction}}

\begin{figure}[pos=htbp]
	\centering
	\includegraphics[width=0.6\linewidth]{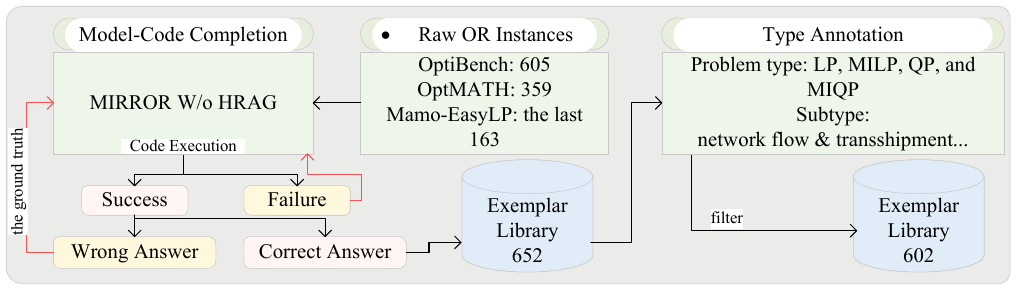}
	\caption{Construction workload of the exemplar library.}
	\vspace{-0.6cm}
	\label{ex}
\end{figure}

The exemplar library $\mathcal{L}$ is constructed through a multi-stage curation pipeline during the preparation phase. Figure~\ref{ex} presents the construction workflow. At this stage, ground-truth verification signals are available for evaluating correctness, including both the executability of the solver code and the optimality of the resulting solution.
We collect 1,127 raw optimization problems from three sources: 605 instances from OptiBench~\citep{yang2025optibench}, 359 instances from ``OptMATH'', and the final 163 instances from ``Mamo-EasyLP''.

Each instance is first processed by MIRROR without \textsc{HRAG} to complete its missing mathematical formulation and executable solver code. The generated model--solver-code pair is then verified by an oracle solver. This completion process is conducted for a fixed number of iterations. If the generated code fails to compile, the revision workflow is activated to correct both the model and the code. If the code compiles successfully but its final answer differs from the ground truth, the ground-truth answer is provided to the agents as a reference, and the model and code are regenerated. This procedure continues until the correct answer is obtained or the maximum number of iterations is reached. Only instances that produce the correct optimal solution are retained, yielding 652 fully specified and verified problem--solution pairs.

We then use the large language model \texttt{qwen-plus} to annotate each verified instance with a high-level problem category, including linear programming (LP), mixed-integer linear programming (MILP),
quadratic programming (QP), and mixed-integer quadratic programming (MIQP), and a fine-grained subproblem type. The subproblem taxonomy comprises blending \& mixing; network flow \& transshipment; multi-period production \& inventory; resource allocation; sequencing \& routing; location \& network design; selection \& knapsack; discrete scheduling \& assignment; covering, packing \& partitioning; risk \& return balancing; error minimization \& fitting; continuous optimal control; discrete risk control; non-linear layout \& assignment; and constrained sparse regression.

\begin{figure}[pos=htbp]
	\centering
	\includegraphics[width=0.5\linewidth]{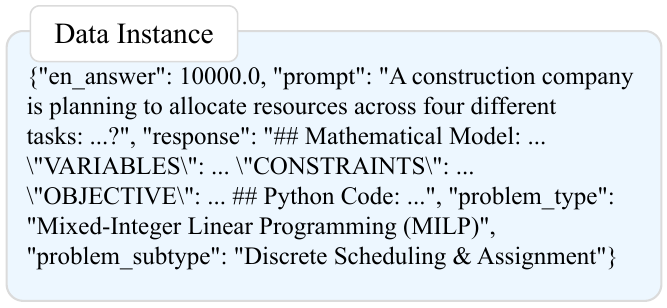}
	\caption{Standard format of a data instance in the exemplar library.}
	\vspace{-0.6cm}
	\label{eng:di}
\end{figure}

Instances whose categories cannot be determined reliably are removed. The remaining 602 instances form the final exemplar library and are encoded using an embedding model to support subsequent retrieval. The standard format of an individual data instance in the exemplar library is shown in Figure~\ref{eng:di}.

\paragraph{\textbf{Hierarchical Retrieval Process}}

\begin{figure}[pos=htbp]
	\centering
	\includegraphics[width=0.6\linewidth]{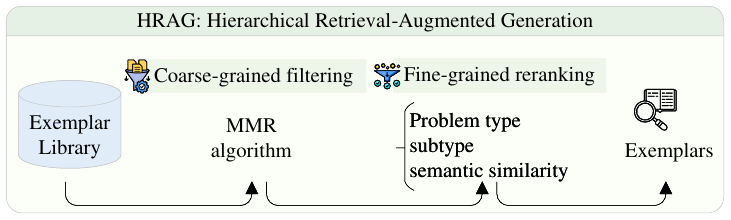}
	\caption{Hierarchical retrieval process.}
	\vspace{-0.6cm}
	\label{hragf}
\end{figure}

The Mathematical Modeling Agent and the Code Generation Agent retrieve their respective reference exemplars through a two-stage process, the detailed procedure of which is illustrated in Figure~\ref{hragf}.

\textbf{1. Coarse-grained filtering:} The embedding model $\Theta_{\text{emb}}$ is combined with the Maximal Marginal Relevance (MMR) algorithm~\citep{adams2022combiningstateoftheartmodelsmaximal} to identify candidate exemplars that are both semantically relevant and diverse.

\textbf{2. Fine-grained reranking:} A large language model reranks the candidate exemplars according to problem-category alignment and deeper semantic similarity, thereby selecting the exemplars most relevant to the target task. The system delivers at most two high-value exemplars to each agent. These are represented as the modeling reference $R_{\text{model}}$ and the code reference $R_{\text{code}}$ for the Mathematical Modeling Agent and the Code Generation Agent, respectively. If no suitable exemplar is identified, the system returns a null signal and activates a stable fallback mechanism.

\subsection{Revision Phase}
\label{eng:revision}

\subsubsection{Iterative Adaptive Revision (\textsc{IAR})}
\label{eng:iar}

\begin{figure*}[pos=htbp]
	\centering
	\includegraphics[width=\linewidth]{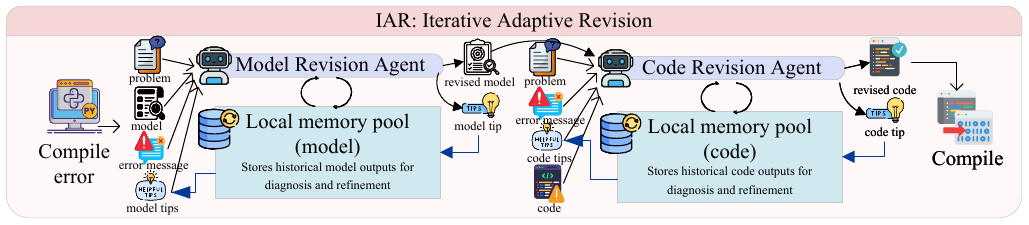}
	\caption{Iterative adaptive revision process.}
	\vspace{-0.6cm}
	\label{iar}
\end{figure*}

The system first performs forward generation by sequentially invoking $f_{\text{param}}$, $f_{\text{adv}}$, $f_{\text{model}}$, and $f_{\text{code}}$, thereby mapping the input task $t$ to executable solver code $c$. If the executor returns a failure, the \textsc{IAR} mechanism is activated. The Mathematical Modeling Agent and the Code Generation Agent then switch roles and operate as the \textbf{Modeling Revision Agent} $\delta_{\text{model}}$ and the \textbf{Code Revision Agent} $\delta_{\text{code}}$, respectively. These agents collaboratively diagnose the root cause of the failure and generate structured revision tips. The Dual Memory mechanism described in Section~\ref{eng:dual-memory} supports error diagnosis and correction throughout this process. The detailed procedure is depicted in Figure~\ref{iar}.
When execution fails, the Modeling Revision Agent first enters the revision process. It retrieves the most recently generated model and available revision tips from the modeling memory pool, which is empty during the initial revision round. Using the retrieved content, the original problem description, and the error message returned by the executor, the agent revises the mathematical model and stores the updated model in the modeling memory pool. It also generates a structured modeling revision tip for use in subsequent revision rounds.

\begin{figure}[pos=htbp]
	\centering
	\begin{subfigure}[b]{0.48\textwidth}
		\centering
		\includegraphics[width=\linewidth]{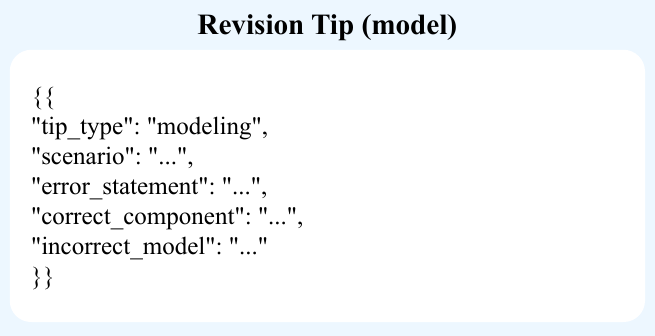}
		\caption{Modeling revision tip.}
		\label{eng:rtm}
	\end{subfigure}
	\hfill
	\begin{subfigure}[b]{0.48\textwidth}
		\centering
		\includegraphics[width=\linewidth]{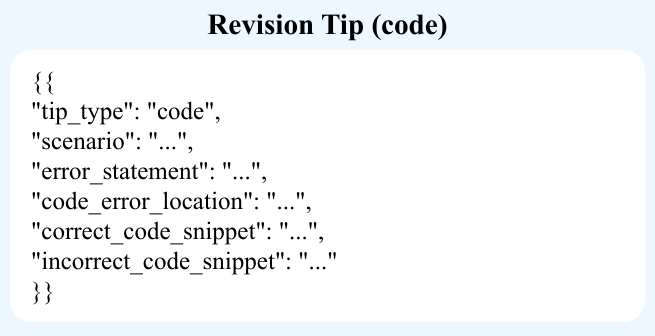}
		\caption{Code revision tip.}
		\label{eng:rtc}
	\end{subfigure}
	\caption{Structured revision tip templates.}
	\vspace{-0.6cm}
	\label{eng:rt}
\end{figure}

After the Modeling Revision Agent completes its work, the Code Revision Agent retrieves the latest complete code and revision tips from the programming memory pool, the most recently revised model from the modeling memory pool, and the error message provided by the executor. Based on this information and the original problem description, it revises the solver implementation and produces new code. A programming-related revision tip is generated simultaneously and stored in the programming memory pool. Thus, both revision agents produce structured tips that are retained in their corresponding local memory pools and reused during subsequent revision rounds. The Modeling Revision Agent produces model-level tips, whereas the Code Revision Agent produces implementation-level tips. The revision tip templates are shown in Figure~\ref{eng:rt}.

The system subsequently executes the revised solver code. The model-and-code refinement process continues iteratively until a valid solution is produced or the maximum number of revision rounds is reached, forming a closed-loop correction mechanism driven by execution feedback.

\subsubsection{Dual Memory}
\label{eng:dual-memory}
Recent automated optimization-modeling systems, such as ORMind, have employed shared memory resources to coordinate specialized components by maintaining a centralized Memory Pool that is updated after each component produces an output~\citep{wang2025ormind}. Although shared access promotes information reuse and coordination, placing heterogeneous artifacts in a single common space can blur the boundary between modeling, programming, and execution-feedback information. As the number of agents increases or their outputs become longer, the stored content becomes increasingly redundant and heterogeneous, introducing irrelevant context and retrieval noise that may interfere with error localization during subsequent revision. To address this issue, MIRROR introduces separate local memory pools for mathematical modeling and solver-code generation and revision. Each local pool stores only the historical outputs, error messages, and revision tips associated with its corresponding task, thereby providing focused context for iterative correction. Together with the global memory pool, this dual-memory design supports both cross-task knowledge reuse and task-specific error diagnosis.

\begin{itemize}
	\item \textbf{Local memory pools:} Each local memory pool is associated with a specific type of agent and stores the relevant historical outputs from the generation and revision stages, including previous models, solver code, error messages, and debugging or revision tips. This information provides focused contextual history for iterative revision. The modeling memory pool and the programming memory pool primarily support their corresponding agents while also enabling the necessary transfer of revised modeling information to the Code Revision Agent.
	\item \textbf{Global memory pool:} The global memory pool aggregates outputs produced by all functional agents across different stages and tasks. It enables shared experience and cross-task knowledge reuse, while supporting continuous optimization at the system level.
\end{itemize}

\subsection{Illustrative Examples}
\label{eng:instance}

\subsubsection{Revision Instance}

\begin{figure}[pos=htbp]
	\centering
	\begin{subfigure}[b]{6.04cm}
		\centering
		\includegraphics[height=5.5cm]{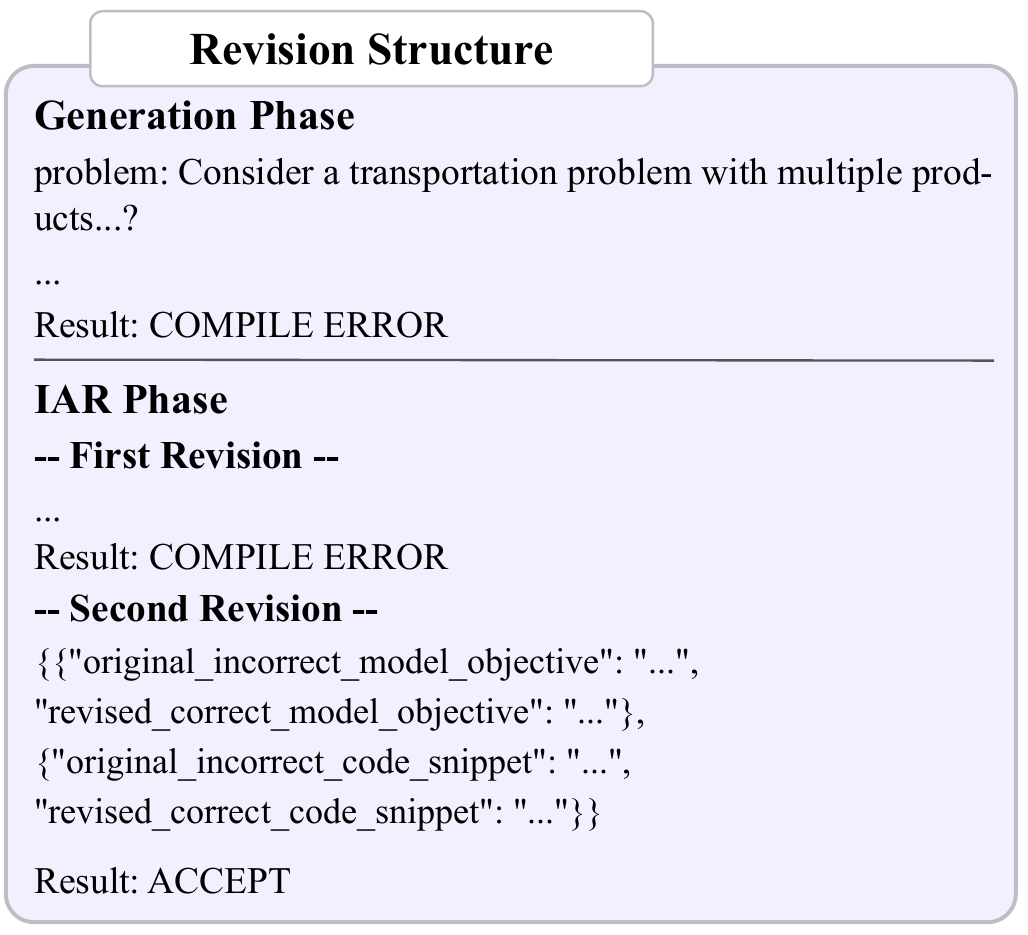}
		\caption{Revision structure.}
		\label{eng:rs}
	\end{subfigure}
	\hspace{0.50cm}
	\begin{subfigure}[b]{8.07cm}
		\centering
		\includegraphics[height=5.5cm]{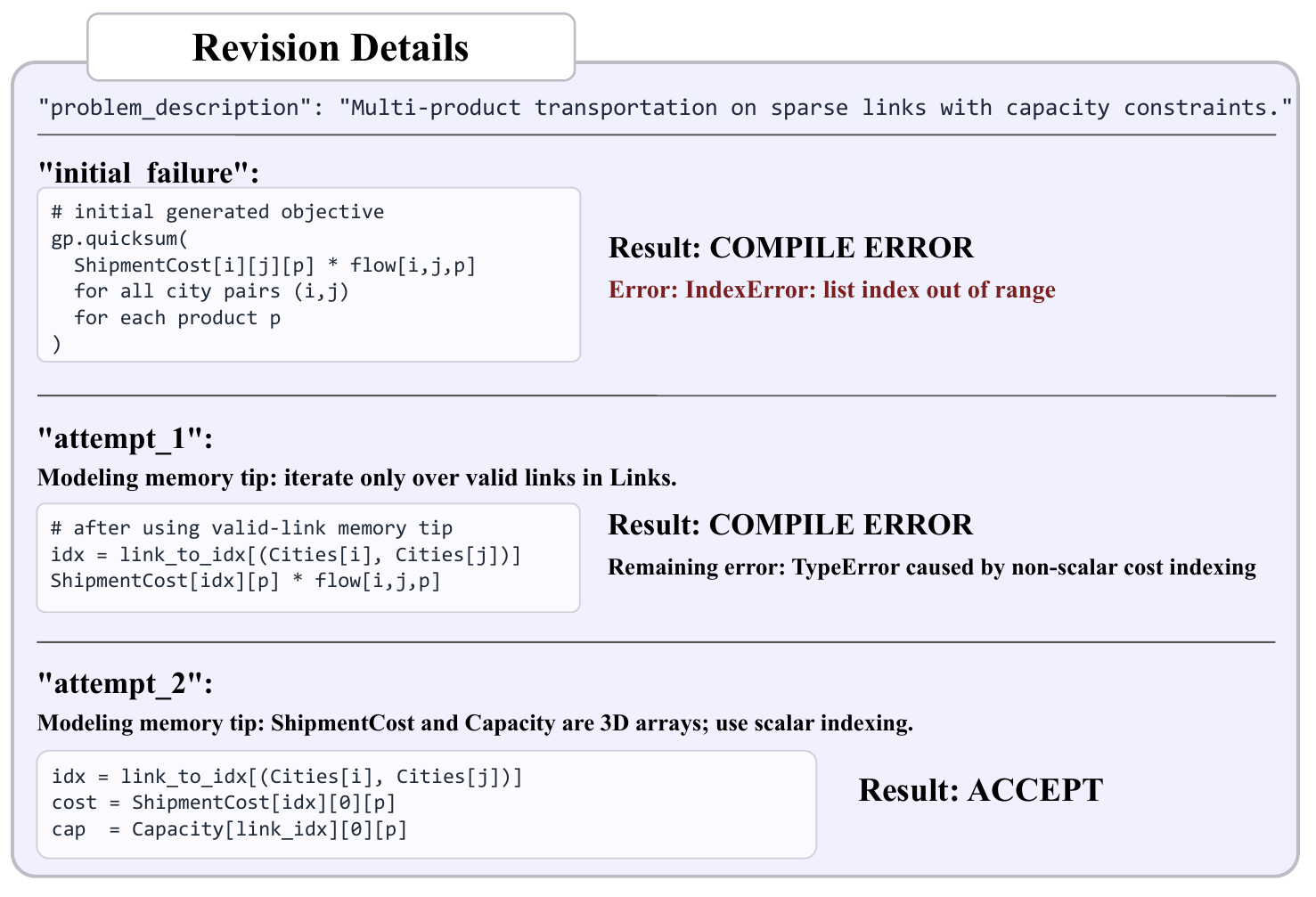}
		\caption{Revision details.}
		\label{eng:rd}
	\end{subfigure}
	\caption{An example of the MIRROR revision process.}
	\vspace{-0.6cm}
	\label{eng:ri}
\end{figure}

Figure~\ref{eng:rs} presents the overall structure of the revision process, while Figure~\ref{eng:rd} illustrates the detailed revision trajectory of a specific problem. In this example, the initially generated code fails to compile, but the problem is successfully solved and the correct answer is obtained after two revision rounds.

During the initial implementation, the Code Generation Agent incorrectly assumes that the transportation network is fully connected and constructs the objective function by traversing the Cartesian product of all city pairs. However, the input data contain only a specified set of sparse directed links. Consequently, the generated solver code attempts to access \texttt{ShipmentCost} entries for non-existent links and raises an \texttt{IndexError} (\texttt{list index out of range}).

In response, the system initiates the first revision round (Attempt 1). The Modeling Revision Agent identifies the mismatch between the mathematical model and the sparse data structure and determines that the objective function must iterate only over the links explicitly listed in \texttt{Links}. The Code Revision Agent accordingly introduces a \texttt{link\_to\_idx} mapping and modifies the implementation to traverse valid links only. Although this revision resolves the out-of-range indexing problem, the revised code raises a \texttt{TypeError} at runtime. The error indicates that a sequence is being multiplied by a Gurobi variable, revealing that the code still applies an incorrect indexing depth to the three-dimensional \texttt{ShipmentCost} array and therefore fails to extract a scalar cost value.

The system then performs a second revision round (Attempt 2) to resolve this subtler data-structure error. MIRROR analyzes the three-dimensional list structure of \texttt{ShipmentCost}, and the Modeling Revision Agent generates a critical memory tip stating that the index expression must precisely match the dimensional organization of the input data. Guided by this tip, the Code Revision Agent corrects the indexing of both the cost and capacity arrays, for example replacing \texttt{ShipmentCost[...][p]} with \texttt{ShipmentCost[...][0][p]}.

After these two feedback-driven revision rounds, the solver code compiles and executes successfully, producing the final result \texttt{ACCEPT}. This example demonstrates that \textsc{IAR} can correct not only surface-level syntax errors but also deeper logical defects and data-structure alignment errors through repeated execution, diagnosis, and revision, thereby improving the correctness and robustness of the final solution.

\subsubsection{Exemplar Structure}

\begin{figure}[pos=htbp]
	\centering
	\begin{subfigure}[b]{0.48\textwidth}
		\centering
		\includegraphics[width=\linewidth]{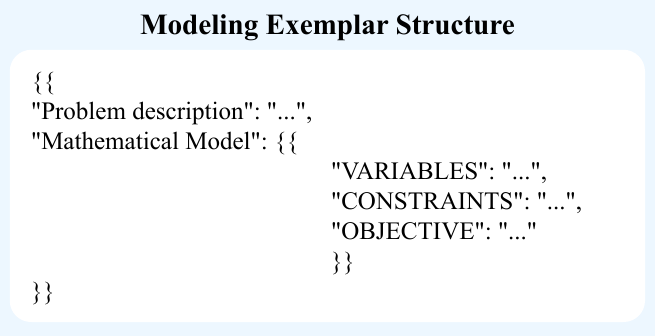}
		\caption{Modeling exemplar structure.}
		\label{eng:mes}
	\end{subfigure}
	\hfill
	\begin{subfigure}[b]{0.48\textwidth}
		\centering
		\includegraphics[width=\linewidth]{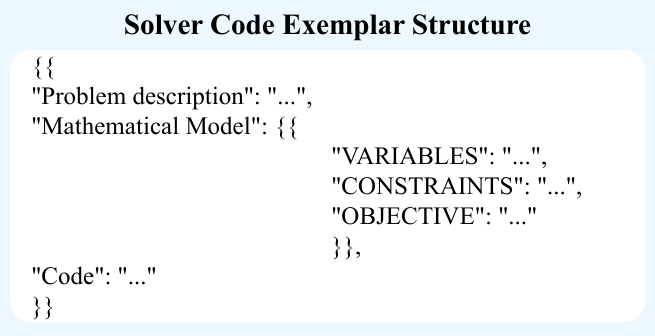}
		\caption{Solver code exemplar structure.}
		\label{eng:sces}
	\end{subfigure}
	\caption{Structures of the retrieved reference exemplars.}
	\vspace{-0.6cm}
	\label{eng:es}
\end{figure}

The reference exemplars ultimately provided to the Mathematical Modeling Agent and the Code Generation Agent use the structures shown in Figure~\ref{eng:es}.

\begin{figure}[pos=htbp]
	\centering
	\begin{subfigure}[t]{0.48\textwidth}
	\vspace{0pt}
		\centering
		\includegraphics[width=\linewidth]{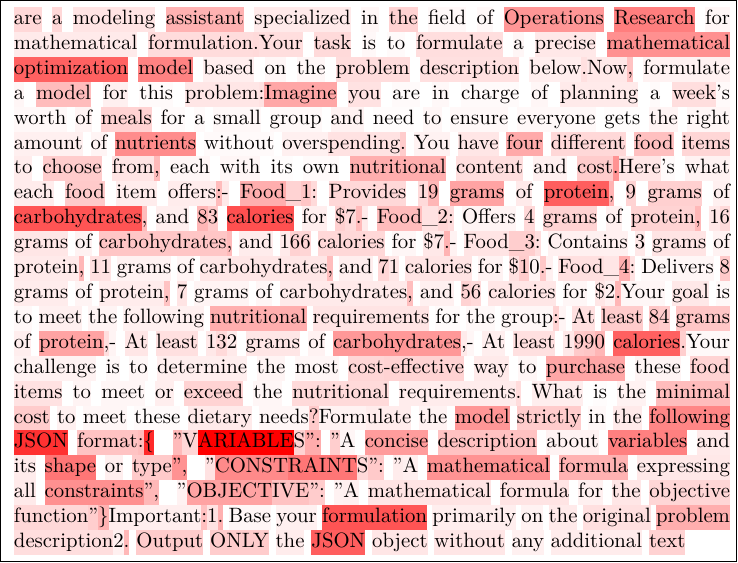}
		\caption{Without exemplars.}
		\label{eng:heatmap-no}
	\end{subfigure}
	\hfill
	\begin{subfigure}[t]{0.48\textwidth}
	\vspace{0pt}
		\centering
		\includegraphics[width=\linewidth]{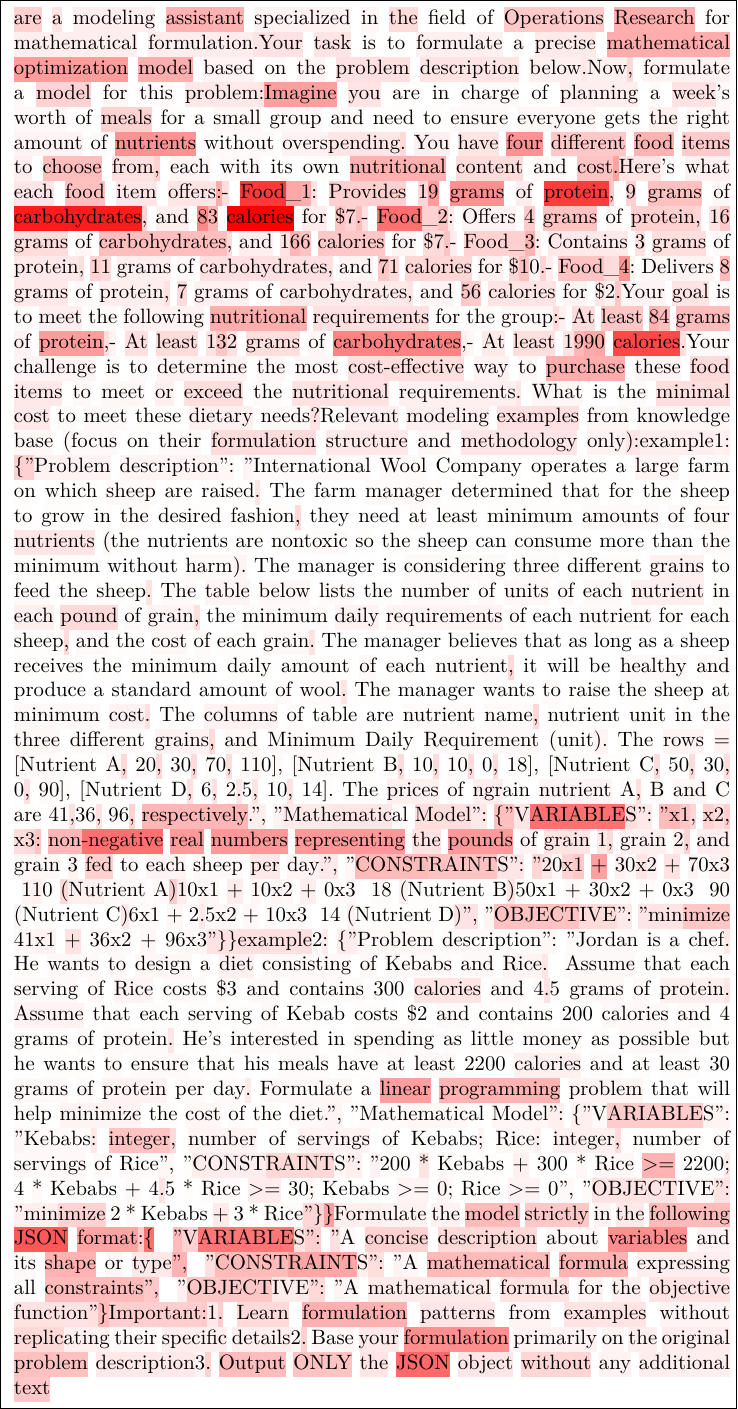}
		\caption{With HRAG exemplars.}
		\label{eng:heatmap-yes}
	\end{subfigure}
	\caption{Comparison of token-level attribution heatmaps.}
	\vspace{-0.6cm}
	\label{eng:heatmap-hrag}
\end{figure}
\subsection{Token-Level Attribution Analysis}
\label{eng:attn}

To investigate how the exemplars retrieved by \textsc{HRAG} influence the framework's output, we employ Attention-aware Layer-wise Relevance Propagation (AttnLRP)~\citep{achtibat2024attnlrp} and follow the token-level relevance analysis method proposed by~\citep{huang2026heuristic}. All experiments are conducted using Qwen2.5-7B-Instruct with the \texttt{lxt} library and 4-bit NormalFloat (NF4) quantization.

We select a problem instance for which the model produces an incorrect answer without \textsc{HRAG} but a correct answer after \textsc{HRAG} is enabled. The analysis focuses on the modeling stage and examines how the retrieved exemplars affect the model's internal relevance flow. In the heatmaps, darker colors indicate that the corresponding input tokens exert a stronger influence on the generated output.

Because decoder-only Transformers employ causal self-attention, the first token in a sequence can accumulate a disproportionately large relevance score. Its key and value representations are attended to by all subsequent positions, and relevance is propagated back to this position from downstream tokens during the Layer-wise Relevance Propagation (LRP) backward pass. The first token consequently acts as a relevance sink and can obscure the contributions of other informative tokens. We therefore exclude it from the visualization to reveal more meaningful relevance patterns in the remainder of the prompt.

We compare the following two prompt conditions:

\begin{itemize}
	\item \textbf{Condition 1 (with HRAG exemplars):} The prompt contains a role description, a task description with the complete problem text, two relevant exemplars retrieved from the exemplar library, and an output-format specification.
	\item \textbf{Condition 2 (without exemplars):} The retrieval results are omitted, and the prompt contains only the role description, problem description, and output-format specification.
\end{itemize}

As shown in Figure~\ref{eng:heatmap-hrag}, without exemplars, generic prompt terms such as ``variables'', ``constraints'', ``objective'', ``optimization'', ``formulation'', and ``JSON'', which do not convey the specific semantics of the target problem, exert a relatively strong influence on the output. After the exemplars are introduced, the model captures structural information such as ``linear programming'' from the retrieved examples, which facilitates construction of the correct mathematical model. Compared with the zero-shot condition, the model also assigns greater relevance to domain-specific terms in the problem description, such as ``calories'' and ``protein''. These results indicate that the retrieved exemplars guide the model toward both the appropriate optimization structure and the task-specific semantic information, providing further evidence of the effectiveness of the proposed \textsc{HRAG} mechanism.

\section{Experiments}
\label{experiment}
In this section, we evaluate MIRROR on five diverse datasets to assess its optimization solving capabilities. 

\subsection{Experimental Setup}
\label{Experimental Setup}
\paragraph{Benchmarks}
We evaluate our method on five standard benchmarks in the domain of optimization modeling: ``NL4Opt''\-~\citep{ramamonjison2023nl4opt}, Mamo~\citep{huang2025llmsmathematicalmodelingbridging}, IndustryOR~\citep{chen2025solverinformed}, and ``ComplexOR''~\citep{xiao2023chain}. From the Mamo benchmark, we specifically utilize two subsets: Mamo-EasyLP and Mamo-ComplexLP. According to the complexity analysis in~\citep{xiao2025survey}, which quantifies problem difficulty based on the number of variables and constraints, NL4Opt, Mamo-EasyLP, and ComplexOR are classified as simple tasks, whereas IndustryOR and Mamo-ComplexLP are categorized as complex tasks. Additionally, the last 163 instances of Mamo-EasyLP are reserved as part of the source data for the exemplar library $\mathcal{L}$, while the remaining 489 instances constitute the test set. 
\paragraph{Baselines}
We compare MIRROR against three categories of methods:

(1) \textbf{Traditional prompting}: Standard Chain-of-Thought (CoT)~\citep{wei2022chain} prompting\- is applied on four models: the backbone model (\texttt{qwen-plus-2025-09-11}), \texttt{DeepSeek-v3}, \texttt{glm-5.1}~\citep{glm5team2026glm5vibecodingagentic} , and \texttt{qwen3-30B} to evaluate conventional inference under different LLMs.

(2) \textbf{Learning-based models}: MiniOpt, LLMOPT, OptMATH, ORLM, and SIRL---methods that fine-tune specialized models on large-scale optimization datasets for end-to-end modeling.

(3) \textbf{Agent-based methods}: OptiMUS, ORMind, OptiTree and Chain-of-Experts (COE), which use multi-agent LLM frameworks with division-of-labor collaboration. All are implemented using the same backbone model as our method for fair comparison.

Additionally, we validate MIRROR’s effectiveness on a smaller model: \texttt{qwen3-30b-a3b-instruct\--2507} (abbreviated as qwen3-30B), comparing CoT prompting against the full MIRROR framework.

\paragraph{Implementation Details}
Except for the small-model ablation, all experiments use \texttt{qwen-plus\--2025-09-11} as the backbone model, with temperature set to $ 0 $ for deterministic and reproducible outputs. Generated code is formatted for the Gurobi solver.

\paragraph{Evaluation Metric}
We adopt \textbf{pass@1}~\citep{chen2021evaluatinglargelanguagemodels} as the primary metric, defined as the proportion of problem instances for which the model generates a correct solution in a single attempt. A prediction $\hat{y}$ is considered correct if it satisfies the relative error tolerance with respect to the ground-truth optimal value $y^*$:
\begin{equation}
	\frac{|y^* - \hat{y}|}{|y^*|} < 10^{-3},
\end{equation}
and to handle the case where $y^* = 0$ , we introduce an absolute error criterion: $|y^* - \hat{y}| < 10^{-1}$.

\subsection{Results Analysis}
\paragraph{Overall Performance}
MIRROR achieves the highest rank among all fully evaluated methods in Table~\ref{tab:results_compact}, setting a new state of the art among multi-agent approaches for end-to-end optimization modeling. Notably, this performance is attained without any task-specific fine-tuning, relying solely on collaborative reasoning, hierarchical retrieval, and iterative adaptive revision.

\begin{table}[pos=htbp]
	\centering
	\setlength{\tabcolsep}{2.3pt}
	\caption{Accuracy (pass@1, \%) of different methods across benchmark datasets.}
	\label{tab:results_compact}
	\begin{tabular*}{\tblwidth}{@{}l l c c c c c c c@{}}
		\toprule
		\textbf{Category} & \textbf{Models / Methods} & \textbf{NL4Opt} & 
		\makecell{\textbf{Mamo}\\\textbf{EasyLP}} & 
		\makecell{\textbf{Mamo}\\\textbf{ComplexLP}} & 
		\textbf{IndustryOR} & \textbf{ComplexOR} & \makecell{\textbf{Macro}\\\textbf{Avg}} & \textbf{Rank} \\
		\midrule
		
		\multirow{4}{*}{\begin{tabular}[c]{@{}l@{}}\textit{Traditional} \\ \textit{prompting}\end{tabular}}
		& Backbone model & 72.24 & 85.07 & 44.83 & 46.00 & 44.44 & 58.52 & 9\\
		&  Deepseek-v3 & 73.88 & 84.66 & 57.14 & 51.00 & 55.56 & 64.45 & 5 \\ 
		& glm-5.1 &77.96 & 89.16 & 56.16 & \textbf{58.00} & 50.00 & 66.26 & 4 \\
		& qwen3-30B &68.57 & 85.28 & 30.54 & 41.00 & 33.33 & 51.74 & 12 \\ 
		\midrule
		\multirow{5}{*}{\begin{tabular}[c]{@{}l@{}}\textit{Learning-based} \\ \textit{methods}\end{tabular}}
		& MiniOpt (14B)\textsuperscript{*} & 92.17 & 90.80 & 33.65 & 27.00 & 61.11 & 60.95 & 8 \\ 
		& LLMOPT (14B)\textsuperscript{*} & 80.28 & 89.53 & 44.08 & 29.00 & 35.29 & 55.64 & 10 \\ 
		& OptMATH (7B)\textsuperscript{*} & 78.70 & 84.20 & 34.12 & 19.00 & 33.33 & 49.87 & 13\\ 
		& ORLM (8B)\textsuperscript{*} & 85.70 & 82.30 & 37.40 & 38.00 & — & — & —\\ 
		& SIRL (32B)\textsuperscript{*} & \textbf{98.00} & \textbf{94.60} & 61.10 & 48.00 & — & — & — \\ 
		
		\midrule
		\multirow{5}{*}{\begin{tabular}[c]{@{}l@{}}\textit{Agent-based} \\ \textit{methods}\end{tabular}}
		& OptiMUS & 57.96 & 85.28 & 45.81 & 51.00 & 27.78 & 53.57 & 11\\ 
		& COE & 84.90 & 87.93 & 57.63 & 55.00 & 55.56 & 68.20 & 3\\ 
		& ORMind & 77.55 & 81.19 & 52.22 & 51.00 & 44.44 & 61.28 & 7\\ 
		& OptiTree & 87.77 & 86.57 & 67.00 & 51.00 & 55.56 & 69.58 & 2\\

		\midrule
		\multirow{2}{*}{\textit{\textbf{Ours}}}    & MIRROR & 86.50 & 87.30 & \textbf{67.50} & 57.00 & \textbf{61.11} & \textbf{71.88} & \textbf{1} \\ 
		& MIRROR (30B) & 82.40 & 86.90 & 52.70 & 53.00 & 44.44 & 63.89 & 6\\ 
		\bottomrule
	\end{tabular*}
	\medskip
	\footnotesize
	All agent-based methods use the default model. Bold denotes the current state-of-the-art (SOTA); ``---'' indicates unreported results; values marked with~\textsuperscript{*} are from other papers: SIRL and ORLM from their original works, and all other learning-based methods from MiniOpt.
\end{table}
\paragraph{Advancement over Agent-based Baselines}
Among existing agent-based methods, OptiTree achieves the highest macro-average accuracy (69.58\%), outperforming OptiMUS (53.57\%), COE (68.20\%), and ORMind (61.28\%). MIRROR surpasses all of them, exceeding OptiTree on four out of five benchmarks and achieving improvements of +6.0\% on IndustryOR and +5.6\% on ComplexOR. This consistent improvement demonstrates that the IAR and HRAG mechanisms capture complex constraints and non-standard problem structures more effectively than previous agent designs.

Table~\ref{tab:inference_cost} reports the token consumption of various agent-based methods on the IndustryOR and ComplexOR datasets. 
\begin{table}[pos=htbp]
	\centering
	\begin{minipage}{0.6\textwidth}
	\caption{Comparison of total token consumption and accuracy across agent-based methods.}
	\label{tab:inference_cost}
	\small
	\begin{tabular*}{\linewidth}{@{\extracolsep{\fill}}lcccc@{}}
		\toprule
		\textbf{Model/Variant} & \multicolumn{2}{c}{\textbf{ComplexOR}} & \multicolumn{2}{c}{\textbf{IndustryOR}} \\
		\cmidrule(lr){2-3} \cmidrule(l){4-5}
		 & \textbf{Tokens} & \textbf{Acc.} & \textbf{Tokens} & \textbf{Acc.} \\
		\midrule
		MIRROR & 209k & 61.11\% & 1068k & 57.00\% \\
		COE & 352k & 55.56\% & 1527k & 55.00\% \\
		ORMind & 106k & 44.44\% & 743k & 51.00\% \\
		OptiMUS & 410k & 27.78\% & 3953k & 51.00\% \\
		OptiTree & 124k & 55.56\% & 732k & 51.00\% \\
		\bottomrule
	\end{tabular*}
	\end{minipage}
\end{table}
From the results, we can observe that: (1) MIRROR consumes fewer tokens than COE and OptiMUS on both datasets; on IndustryOR, its token usage is 70\% of COE's and 27\% of OptiMUS's; (2) Although ORMind and OptiTree consume fewer total tokens than MIRROR on both datasets, MIRROR achieves accuracy improvements over ORMind by 6 percentage points on IndustryOR and by 16.67 percentage points on ComplexOR, with the advantage on ComplexOR being particularly pronounced. Moreover, MIRROR also outperforms OptiTree by 6 percentage points on IndustryOR and by 5.55 percentage points on ComplexOR. In many practical industrial problems, accuracy is of paramount importance, as erroneous answers can lead to severe consequences.

\paragraph{Superiority to Learning-based Models}
MIRROR requires no task-specific training yet outperforms all learning-based baselines in terms of average performance: MiniOpt (14B), LLMOPT (14B), OptMATH (7B), ORLM (8B), and SIRL (32B). On the challenging Mamo-ComplexLP and IndustryOR benchmarks, it achieves 67.50\% and 57.00\% accuracy, respectively---surpassing the strongest prior model, SIRL (32B), by 6.40 and 9.00 percentage points. This shows that structured agent collaboration can excel at complex optimization modeling without large-scale supervised fine-tuning.

\paragraph{Advantage Over Traditional Prompting Approach}
MIRROR, using the backbone model, achieves a macro-average score of 71.88\%, outperforming Chain-of-Thought (CoT) prompting on the same model (58.52\%) by 13.4 points. It also exceeds CoT applied to the stronger \texttt{DeepSeek-v3} model (64.45\%) and \texttt{glm-5.1} model (66.26\%).

\paragraph{Effective Transfer to Small Model}
When applied to qwen3-30B, MIRROR boosts macro-average accuracy from 51.74\% (achieved by CoT) to 63.89\%---a 12.15-point improvement without fine-tuning, confirming its plug-and-play utility for accessible models.

\subsection{Ablation Study}
We conduct an ablation study on the five datasets: NL4Opt, Mamo-EasyLP, Mamo-ComplexLP, IndustryOR, and ComplexOR. We evaluate four variants: the full MIRROR (with both IAR and HRAG), and three ablated versions---removing only IAR, only HRAG, or both. As shown in Table~\ref{tab:ablation}, the full model achieves the highest macro-average accuracy of 71.88\%, outperforming the w/o HRAG variant (67.90\%), w/o IAR (71.14\%), and w/o Both (65.82\%). This confirms that both mechanisms contribute to performance, with HRAG yielding a larger gain.
\begin{table}[pos=htbp]
	\centering
	\caption{Ablation study of MIRROR components (Accuracy (pass@1, \%)).}
	\label{tab:ablation}
	\begin{tabular*}{\tblwidth}{@{\extracolsep{\fill}}l c c c c c c@{}}
		\toprule
		\textbf{Variant} & \textbf{NL4Opt} & \textbf{Mamo EasyLP} & \textbf{Mamo ComplexLP} & \textbf{IndustryOR} & \textbf{ComplexOR} & \textbf{Macro Avg}\\
		\midrule
		MIRROR & \textbf{86.50} & \textbf{87.30} & \textbf{67.50} & \textbf{57.00} & \textbf{61.11} & \textbf{71.88} \\
		W/o IAR & 85.70 & 86.90 & 67.00 & 55.00 & 61.11 & 71.14 \\
		W/o HRAG & 85.30 & 86.70 & 63.50 & 54.00 & 50.00 & 67.90 \\
		W/o Both & 84.10 & 86.50 & 62.05 & 52.00 & 44.44 & 65.82 \\
		\bottomrule
	\end{tabular*}
\end{table}

We further decompose errors into wrong answer rate (executable solver code that generates an incorrect numerical result) and compile error rate (syntactically invalid solver code that fails to execute), as these reflect distinct failure modes in optimization modeling: the former indicates flawed reasoning, while the latter reveals structural or grammatical mistakes in the generated solver code. Figure~\ref{fig:ablation_study} visualizes these two error types across configurations. The light-colored regions in Figure~\ref{fig:answer_error} represent the compile error rate. Since the wrong answer rates are significantly higher than the compile error rates in most cases across the datasets and mechanisms, the light-colored regions for the former are omitted in Figure~\ref{fig:compile_error} to ensure a clearer visualization of the latter.
{
	\captionsetup[subfigure]{labelfont=normalfont, font=normalfont, labelformat=empty}
	\begin{figure}[pos=htbp]
		\centering
		
		\begin{subfigure}[b]{0.48\linewidth}
			\centering
			\includegraphics[width=\textwidth]{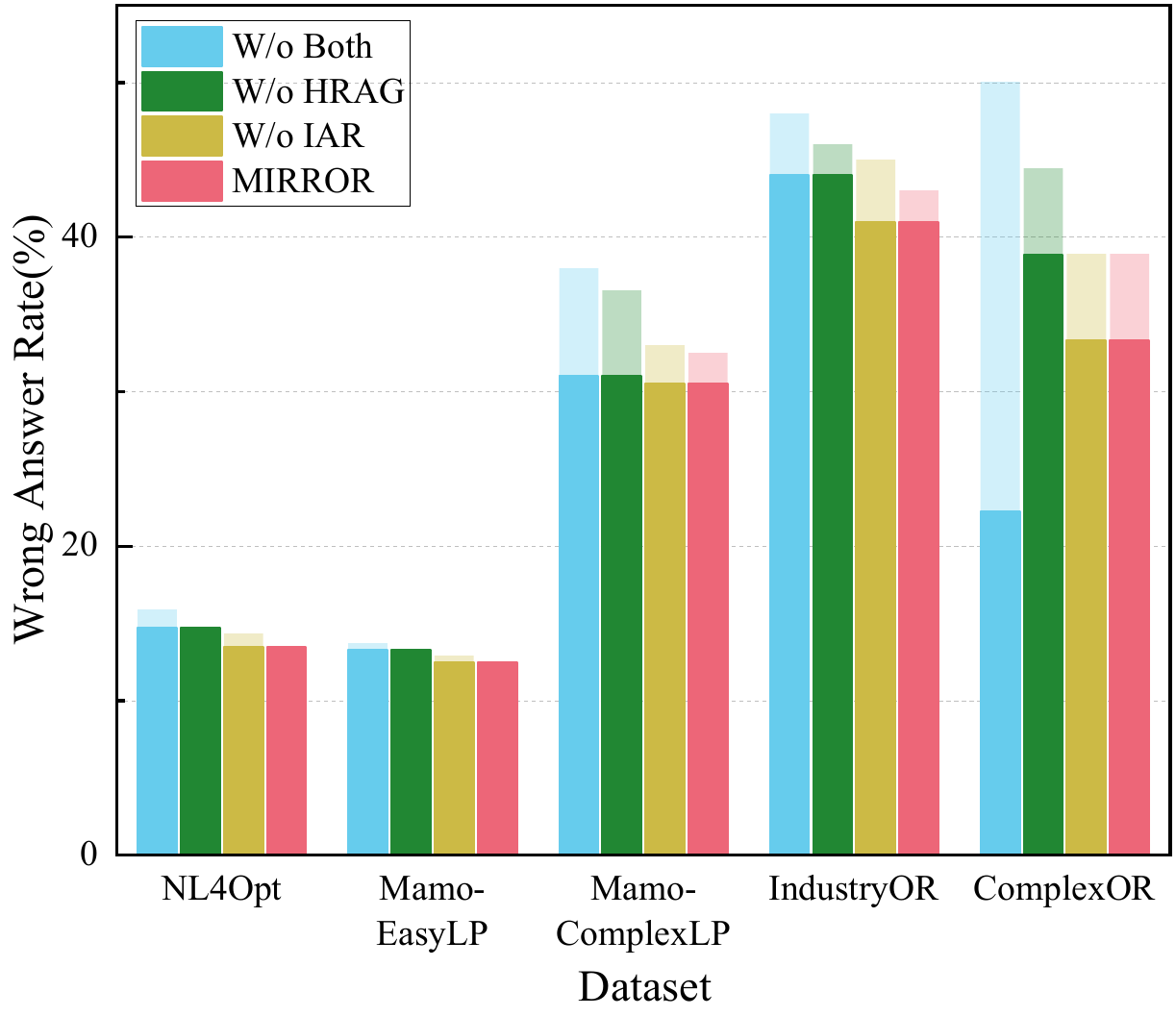}
			\caption{(a) Wrong answer rate across datasets and methods.}
			\label{fig:answer_error}
		\end{subfigure}
		\hfill
		\begin{subfigure}[b]{0.48\linewidth}
			\centering
			\includegraphics[width=\textwidth]{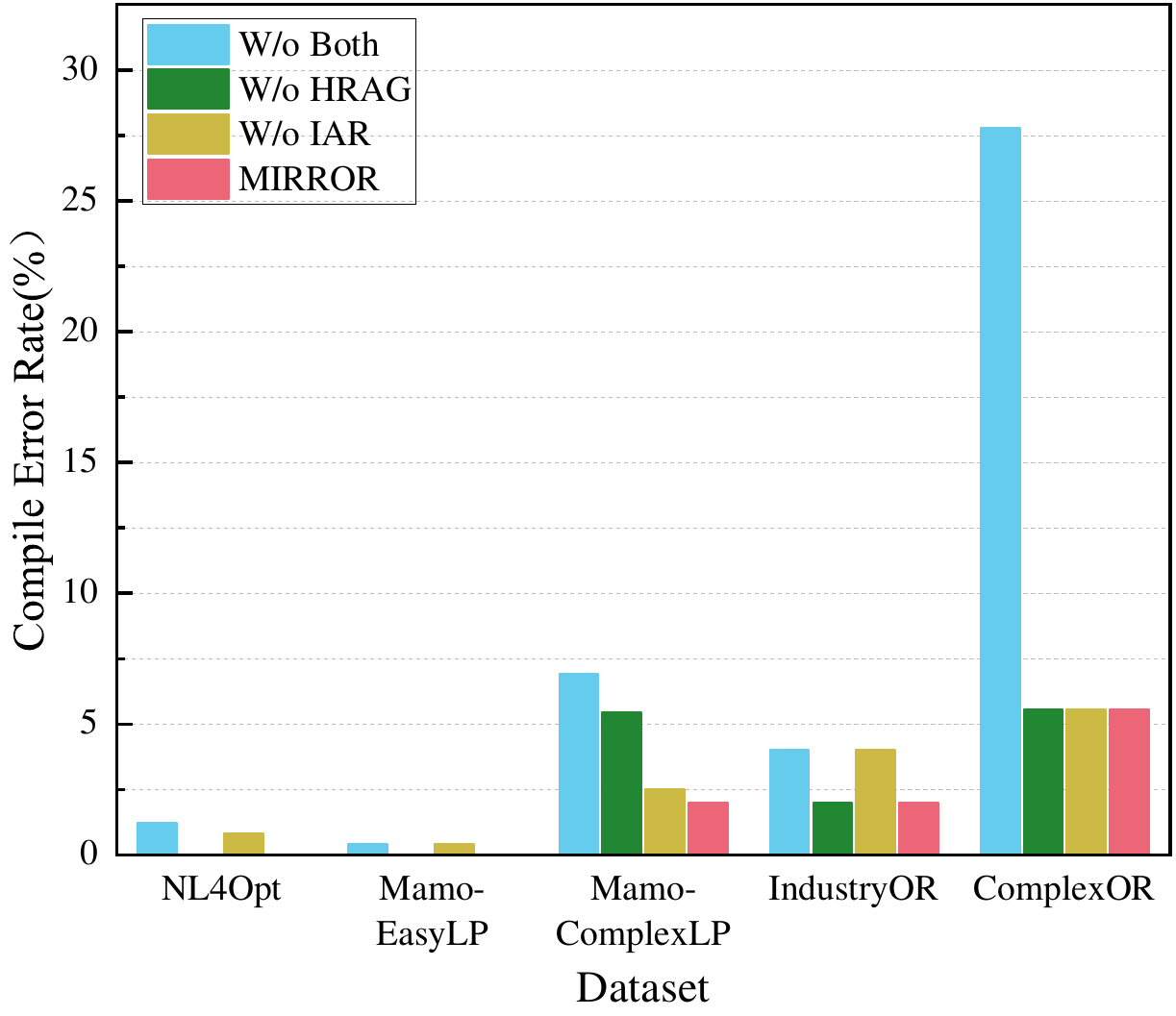}
			\caption{(b) Compile error rate across datasets and methods.}
			\label{fig:compile_error}
		\end{subfigure}
		\caption{
			Ablation study of MIRROR.
		}
		\label{fig:ablation_study}
		\vspace{-0.6cm}
	\end{figure}
}
\paragraph{For Wrong Answer Rate} all datasets except ComplexOR show a consistent decrease or remain stable from the w/o Both configuration to single-mechanism variants and finally to the full model, indicating that both mechanisms help produce correct solutions. Notably, removing HRAG leads to a larger increase in wrong answers than removing IAR; for example, on IndustryOR, the wrong answer rate is 3.00\% higher when only HRAG is removed compared to when only IAR is removed. This suggests that HRAG enhances both problem modeling and solver code generation through its two-stage retrieval strategy, thereby improving the accuracy of the final results.

\paragraph{For Compile Error Rate} the opposite trend is observed: on NL4Opt, Mamo-EasyLP, and IndustryOR, the w/o IAR variant incurs higher compile errors than w/o HRAG, demonstrating that the IAR mechanism’s closed-loop iterative adaptive revision process effectively identifies and resolves compilation failures. Averaged across all datasets, the compile error rate decreases from 8.06\% in the w/o Both setting to 2.60\% with IAR only and 2.65\% with HRAG only, and further declines to 1.91\% when both mechanisms are used. This highlights their complementary contributions to generating valid and executable optimization models and solver code.

\paragraph{Agent Ablation Study}\label{aba}
We perform a more fine-grained ablation study to analyze the specific contributions of each agent. We conduct experiments on representative datasets as follows. Since the Code Generation Agent is responsible for generating the final code and the code is required for compilation, we only analyzed three agents during the generation process: the Parameter Extraction Agent (P), the Modeling Advisor Agent (A), and the Mathematical Modeling Agent (M) (abbreviated as P, A, and M respectively in the experimental table). The ablation study selects two representative test sets, IndustryOR and ComplexOR. The results are shown in Table~\ref{tab:ablation_study}.
\begin{table}[pos=htbp]
	\centering
	\begin{minipage}{0.85\textwidth}
		\caption{Ablation study on IndustryOR and ComplexOR datasets.}
		\label{tab:ablation_study}
		\begin{tabular*}{\linewidth}{@{\extracolsep{\fill}}llccc@{}}
			\toprule
			\textbf{Dataset} & \textbf{Configuration} & \textbf{Accuracy} & \textbf{Wrong Answer Rate} & \textbf{Compile Error Rate} \\
			\midrule
			\textbf{IndustryOR} & Full & \textbf{52.00\%} & 44.00\% & \textbf{4.00\%} \\
			& W/o P & 51.00\% & 42.00\% & 7.00\% \\
			& W/o A & 47.00\% & 46.00\% & 7.00\% \\
			& W/o M & 50.00\% & 42.00\% & 8.00\% \\
			\midrule
			\textbf{ComplexOR} & Full & \textbf{44.44\%} & 22.22\% & \textbf{27.78\%} \\
			& W/o P & 38.89\% & 22.22\% & 38.89\% \\
			& W/o A & 38.89\% & 22.22\% & 38.89\% \\
			& W/o M & 38.89\% & 27.78\% & 33.33\% \\
			\bottomrule
		\end{tabular*}
	\end{minipage}
\end{table}

Parameter Extraction Agent (P): After removing P, the compile error rate on ComplexOR increases from 27.78\% to 38.89\% (↑11.11\%), and on IndustryOR from 4.00\% to 7.00\% (↑3.00\%). Removing P leads to an increase in compile error rates on both datasets, indicating that P makes a universal contribution to reducing compilation errors.

Modeling Advisor Agent (A): After removing A, the accuracy on IndustryOR decreases by 5\% (52\% → 47\%), with the wrong answer rate increasing by 2\% (44\% → 46\%); on ComplexOR, accuracy decreases by 5.55\% (44.44\% → 38.89\%), while the wrong answer rate remains unchanged. Removing A leads to a decrease in accuracy on both datasets, indicating that A plays an important role in improving final solution accuracy.

Mathematical Modeling Agent (M): After removing M, the compile error rate on ComplexOR increases from 27.78\% to 33.33\% (↑5.55\%), with accuracy decreasing by 5.55\% (44.44\% → 38.89\%) and wrong answer rate increasing by 5.56\% (22.22\% → 27.78\%). A consistent performance degradation is also observed on the IndustryOR dataset. Removing M leads to negative impacts on either compilability or solution correctness across both datasets, indicating that M plays a key role in ensuring both code compilability and mathematical fidelity.

In summary, removing any of the three agents P, A, or M leads to negative changes in at least one metric, verifying the necessity of all three agents within the framework.
\section{Conclusion}
We present MIRROR, a training-free multi-agent framework for automated operations research (OR) modeling that bridges the gap between natural language problem descriptions and formal mathematical models with executable solver code. By integrating Hierarchical Retrieval-Augmented Generation (HRAG) for dynamic exemplar retrieval and an Iterative Adaptive Revision (IAR) mechanism for execution-driven self-correction, MIRROR effectively mitigates the hallucination and fragility of general-purpose large language models in specialized OR tasks. Experiments show that MIRROR achieves strong performance across diverse OR modeling benchmarks, attaining state-of-the-art results on complex industrial datasets and significantly boosting the capabilities of small open-source language models---without any task-specific training. This work paves the way toward accessible, reliable, and scalable AI-assisted decision-making for real-world operations research applications.
\printcredits

\bibliographystyle{cas-model2-names}

\bibliography{mirror}

@inproceedings{ahmaditeshnizi2024optimus,
  title={{OptiMUS}: Scalable optimization modeling with ({MI}){LP} solvers and large language models},
  author={Ahmaditeshnizi, A. and Gao, W. and Udell, M.},
  booktitle={International Conference on Machine Learning},
  pages={577--596},
  year={2024},
  publisher={PMLR},
}

@inproceedings{
liu2026optitree,
title={{OptiTree}: Hierarchical Thoughts Generation with Tree Search for {LLM} Optimization Modeling},
author={Liu, H. and Wang, J. and Cai, Y. and Han, X. and Kuang, Y. and Hao, J.},
booktitle={The Thirty-ninth Annual Conference on Neural Information Processing Systems},
year={2026},
url={https://openreview.net/forum?id=Ej20yjWMCj}
}

@misc{
thind2026optimai,
title={Optim{AI}: Optimization from Natural Language Using {LLM}-Powered {AI} Agents},
author={Thind, R. and Sun, Y. and Liang, L. and Yang, H.},
year={2026},
url={https://openreview.net/forum?id=JtgZkVdAIP}
}

@article{cannas2024artificial,
  title={Artificial intelligence in supply chain and operations management: A multiple case study research},
  author={Cannas, V. G. and Ciano, M. P. and Saltalamacchia, M. and Secchi, R.},
  journal={International Journal of Production Research},
  volume={62},
  number={9},
  pages={3333--3360},
  year={2024},
  publisher={Taylor \& Francis}
}

@article{guo2025deepseek,
  title={Deep{Seek}-{R}1 incentivizes reasoning in {LLM}s through reinforcement learning},
  author={Guo, D. and Yang, D. and Zhang, H. and Song, J. and Wang, P. and Zhu, Q. and Xu, R. and Zhang, R. and Ma, S. and Bi, X. and others},
  journal={Nature},
  volume={645},
  number={8081},
  pages={633--638},
  year={2025},
  publisher={Nature Publishing Group}
}

@article{achtibat2024attnlrp,
  title={Attnlrp: attention-aware layer-wise relevance propagation for transformers},
  author={Achtibat, R. and Hatefi, S. M. V. and Dreyer, M. and Jain, A. and Wiegand, T. and Lapuschkin, S. and Samek, W.},
  journal={arXiv preprint arXiv:2402.05602},
  year={2024}
}

@article{huang2026heuristic,
  title={From Heuristic Selection to Automated Algorithm Design: LLMs Benefit from Strong Priors},
  author={Huang, Q. and Ye, F. and Shahane, A. and B{\"a}ck, T. and van Stein, N.},
  journal={arXiv preprint arXiv:2603.02792},
  year={2026}
}

@misc{glm5team2026glm5vibecodingagentic,
      title={GLM-5: from Vibe Coding to Agentic Engineering},
      author={GLM-5-Team},
      year={2026},
      eprint={2602.15763},
      archivePrefix={arXiv},
      primaryClass={cs.LG},
      url={https://arxiv.org/abs/2602.15763},
}

@misc{gurobi2024,
  title={{Gurobi} Optimizer Reference Manual},
  author={{Gurobi Optimization, LLC}},
  year={2024},
  url={https://www.gurobi.com},
  note={Reference Manual}
}

@article{huang2025orlm,
  title={{ORLM}: A customizable framework in training large models for automated optimization modeling},
  author={Huang, C. and Tang, Z. and Hu, S. and Jiang, R. and Zheng, X. and Ge, D. and Wang, B. and Wang, Z.},
  journal={Operations Research},
  year={2025},
  publisher={INFORMS}
}

@inproceedings{huang2024knowledgeawarecodegenerationlarge,
  author = {Huang, T. and Sun, Z. and Jin, Z. and Li, G. and Lyu, C.},
  booktitle = {2024 IEEE/ACM 32nd International Conference on Program Comprehension (ICPC)},
  title = {Knowledge-Aware Code Generation with Large Language Models},
  year = {2024},
  pages = {52--63},
  publisher = {IEEE Computer Society},
  month = apr
}

@inproceedings{ramamonjison2023nl4opt,
  title={{NL4Opt} competition: Formulating optimization problems based on their natural language descriptions},
  author={Ramamonjison, R. and Yu, T. and Li, R. and Li, H. and Carenini, G. and Ghaddar, B. and He, S. and Mostajabdaveh, M. and Banitalebi-Dehkordi, A. and Zhou, Z. and others},
  booktitle={NeurIPS 2022 Competition Track},
  pages={189--203},
  year={2023},
  publisher={PMLR}
}

@article{ren2025deepseekproverv2advancingformalmathematical,
  title={{DeepSeek}-{Prover}-{V}2: Advancing formal mathematical reasoning via reinforcement learning for subgoal decomposition},
  author={Ren, Z. Z. and Shao, Z. and Song, J. and Xin, H. and Wang, H. and Zhao, W. and Zhang, L. and Fu, Z. and Zhu, Q. and Yang, D. and others},
  journal={arXiv preprint arXiv:2504.21801},
  year={2025}
}

@inproceedings{pmlr-v235-wang24h,
  title={Executable code actions elicit better {LLM} agents},
  author={Wang, X. and Chen, Y. and Yuan, L. and Zhang, Y. and Li, Y. and Peng, H. and Ji, H.},
  booktitle={Proceedings of the 41st International Conference on Machine Learning},
  pages={50208--50232},
  year={2024},
  volume={235},
  series={Proceedings of Machine Learning Research},
  publisher={PMLR}
}

@inproceedings{jimenez2024swebench,
  title={{SWE}-bench: Can language models resolve real-world {GitHub} issues?},
  author={Jimenez, C. E. and Yang, J. and Wettig, A. and Yao, S. and Pei, K. and Press, O. and Narasimhan, K.},
  booktitle={International Conference on Learning Representations},
  year={2024}
}

@inproceedings{xiao2023chain,
  title={Chain-of-experts: When {LLM}s meet complex operations research problems},
  author={Xiao, Z. and Zhang, D. and Wu, Y. and Xu, L. and Wang, Y. J. and Han, X. and Fu, X. and Zhong, T. and Zeng, J. and Song, M. and others},
  booktitle={The Twelfth International Conference on Learning Representations},
  year={2023}
}

@article{wei2022chain,
  title={Chain-of-thought prompting elicits reasoning in large language models},
  author={Wei, J. and Wang, X. and Schuurmans, D. and Bosma, M. and Xia, F. and Chi, E. and Le, Q. V. and Zhou, D. and others},
  journal={Advances in Neural Information Processing Systems},
  volume={35},
  pages={24824--24837},
  year={2022}
}

@article{yang2025qwen3,
  title={{Qwen}3 technical report},
  author={Yang, A. and Li, A. and Yang, B. and Zhang, B. and Hui, B. and Zheng, B. and Yu, B. and Gao, C. and Huang, C. and Lv, C. and others},
  journal={arXiv preprint arXiv:2505.09388},
  year={2025}
}

@article{zhang2025or,
  title={{OR}-{LLM}-{Agent}: Automating modeling and solving of operations research optimization problem with reasoning large language model},
  author={Zhang, B. and Luo, P.},
  journal={arXiv preprint arXiv:2503.10009},
  year={2025}
}

@article{wang2025ormind,
  title={{ORMind}: A cognitive-inspired end-to-end reasoning framework for operations research},
  author={Wang, Z. and Chen, B. and Huang, Y. and Cao, Q. and He, M. and Fan, J. and Liang, X.},
  journal={arXiv preprint arXiv:2506.01326},
  year={2025}
}

@inproceedings{
liu2022good,
title={What Makes Good In-Context Examples for {GPT}-3?},
author={Liu, J. and Shen, D. and Zhang, Y. and Dolan, B. and Carin, L. and Chen, W.},
booktitle={Proceedings of the 3rd Workshop on Knowledge Extraction and Integration for Deep Learning Architectures},
year={2022},
pages={100--114},
url={https://aclanthology.org/2022.deelio-1.10/}
}

@article{wu2025step,
  title={Step-{Opt}: Boosting optimization modeling in {LLM}s through iterative data synthesis and structured validation},
  author={Wu, Y. and Zhang, Y. and Wu, Y. and Wang, Y. and Zhang, J. and Cheng, J.},
  journal={arXiv preprint arXiv:2506.17637},
  year={2025}
}

@article{lu2025optmath,
  title={Opt{MATH}: A scalable bidirectional data synthesis framework for optimization modeling},
  author={Lu, H. and Xie, Z. and Wu, Y. and Ren, C. and Chen, Y. and Wen, Z.},
  journal={arXiv preprint arXiv:2502.11102},
  year={2025}
}

@misc{
anonymous2026miniopt,
title={MiniOpt: Reasoning to Model and Solve General Optimization Problems with Limited Resources},
author={Di, Z. and Zhao, K. and Shu, X. and Wen, Y. and Shi, Q. and Qian, H. and Li, B. and Lu, X. and Wang, X. and Zhou, J. and Tang, K. and Yu, Y.},
year={2026},
url={https://openreview.net/forum?id=nIWCUVJ6OU}
}

@article{chen2021evaluatinglargelanguagemodels,
  title={Evaluating large language models trained on code},
  author={Chen, M. and Tworek, J. and Jun, H. and Yuan, Q. and {de O. Pinto}, H. P. and Kaplan, J. and Edwards, H. and Burda, Y. and Joseph, N. and Brockman, G. and others},
  journal={arXiv preprint arXiv:2107.03374},
  year={2021}
}

@inproceedings{
chen2025solverinformed,
title={Solver-Informed {RL}: Grounding Large Language Models for Authentic Optimization Modeling},
author={Chen, Y. and Xia, J. and Shao, S. and Ge, D. and Ye, Y.},
booktitle={The Thirty-ninth Annual Conference on Neural Information Processing Systems},
year={2025},
url={https://openreview.net/forum?id=80L235oVBe}
}

@article{huang2025llmsmathematicalmodelingbridging,
  title={Large language models for mathematical modeling: Towards bridging the gap between natural and mathematical languages},
  author={Huang, X. and Shen, Q. and Hu, Y. and Gao, A. and Wang, B.},
  journal={arXiv preprint arXiv:2405.13144},
  year={2025}
}

@inproceedings{jiang2024llmopt,
  title={{LLMOPT}: Learning to define and solve general optimization problems from scratch},
  author={Jiang, C. and Shu, X. and Qian, H. and Lu, X. and Zhou, J. and Zhou, A. and Yu, Y.},
  booktitle={International Conference on Learning Representations},
  year={2025},
  pages={101580--101606},
}

@article{novikov2025alphaevolvecodingagentscientific,
  title={Alpha{Evolve}: A coding agent for scientific and algorithmic discovery},
  author={Novikov, A. and Vũ, N. and Eisenberger, M. and Dupont, E. and Huang, P.-S. and Wagner, A. Z. and Shirobokov, S. and Kozlovskii, B. and Ruiz, F. J. R. and Mehrabian, A. and others},
  journal={arXiv preprint arXiv:2506.13131},
  year={2025}
}

@article{ge2024cardinaloptimizercoptuser,
  title={Cardinal optimizer ({COPT}) user guide},
  author={Ge, D. and Huangfu, Q. and Wang, Z. and Wu, J. and Ye, Y.},
  journal={arXiv preprint arXiv:2208.14314},
  year={2024}
}

@article{yang2024qwen25mathtechnicalreportmathematical,
  title={{Qwen}2.5-{Math} technical report: Toward mathematical expert model via self-improvement},
  author={Yang, A. and Zhang, B. and Hui, B. and Gao, B. and Yu, B. and Li, C. and Liu, D. and Tu, J. and Zhou, J. and Lin, J. and others},
  journal={arXiv preprint arXiv:2409.12122},
  year={2024}
}

@inproceedings{
liu2025mmagent,
title={{MM}-Agent: {LLM} as Agents for Real-world Mathematical Modeling Problem},
author={Liu, F. and Yang, Z.-R. and Liu, C. and Song, T. and Gao, X. and Liu, H.},
booktitle={2nd AI for Math Workshop @ ICML 2025},
year={2025},
url={https://openreview.net/forum?id=QyKBf7X98d}
}

@inproceedings{xiao2025survey,
author = {Xiao, Z. and Xie, J. and Xu, L. and Guan, S. and Zhu, J. and Han, X. and Fu, X. and Yu, W. and Wu, H. and Shi, W. and Kang, Q. and Duan, J. and Zhong, T. and Yuan, M. and Zeng, J. and Wang, Y. and Chen, G. and Zhang, D.},
title = {A survey of optimization modeling meets {LLMs}: Progress and future directions},
booktitle = {Proceedings of the Thirty-Fourth International Joint Conference on Artificial Intelligence (IJCAI-25)},
year = {2025},
pages = {10742--10750},
doi = {10.24963/ijcai.2025/1192}
}

@article{deepseekai2025deepseekv3,
title = {{DeepSeek-V3} Technical Report},
author = {{DeepSeek-AI} and Liu, A. and Feng, B. and Xue, B. and Wang, B. and Wu, B. and Lu, C. and Zhao, C. and Deng, C. and Zhang, C. and Ruan, C. and Dai, D. and Guo, D. and Yang, D. and Chen, D. and Ji, D. and Li, E. and Lin, F. and Dai, F. and ... and Pan, Z.},
journal = {arXiv preprint arXiv:2412.19437},
year = {2025},
eprint = {2412.19437},
primaryClass = {cs.CL}
}

@article{tang2025calm,
title={{CALM} Before the {STORM}: Unlocking Native Reasoning for Optimization Modeling},
author={Tang, Z. and Ye, Z. and Huang, C. and Huang, X. and Li, C. and Li, S. and Chen, G. and Yan, M. and Wang, Z. and Zha, H. and others},
journal={arXiv preprint arXiv:2510.04204},
year={2025}
}

@article{zhou2025steporlm,
title={{StepORLM}: A self-evolving framework with generative process supervision for operations research language models},
author={Zhou, C. and Xu, T. and Lin, J. and Ge, D.}, journal={arXiv preprint arXiv:2509.22558},
year={2025}
}

@article{ding2025or,
title={{OR}-{R1}: Automating Modeling and Solving of Operations Research Optimization Problem via Test-Time Reinforcement Learning},
author={Ding, Z. and Tan, Z. and Zhang, J. and Chen, T.}, journal={arXiv preprint arXiv:2511.09092},
year={2025}
}

@inproceedings{yang2025optibench,
author = {Yang, Z. and Wang, Y. and Huang, Y. and Guo, Z. and Shi, S. and Han, X. and Feng, L. and Song, L. and Liang, X. and Tang, J.},
title = {OptiBench Meets ReSocratic: Measure and Improve LLMs for Optimization Modeling},
booktitle = {International Conference on Learning Representations},
year = {2025},
pages = {24726--24759}
}

@article{openai2024gpt4technicalreport,
  title={{GPT}-4 technical report},
  author={OpenAI and Achiam, J. and Adler, S. and Agarwal, S. and Ahmad, L. and Akkaya, I. and others},
  journal={arXiv preprint arXiv:2303.08774},
  year={2024}
}

@article{openai2024openaio1card,
  title={{OpenAI} o1 system card},
  author={OpenAI and Jaech, A. and Kalai, A. and Lerer, A. and Richardson, A. and {El-Kishky}, A. and others},
  journal={arXiv preprint arXiv:2412.16720},
  year={2024}
}

@article{liang2026llmlargescaleoptimizationmodel,
  title={{LLM} for Large-Scale Optimization Model Auto-Formulation: Bridging Flexibility and Standardization via Agentic Workflow},
  author={Liang, K. and Lu, Y. and Mao, J. and Sun, S. and Yang, C. and Zeng, C. and Jin, X. and Qin, H. and Zhu, R. and Teo, C.-P.},
  journal={arXiv preprint arXiv:2601.09635},
  year={2026}
}

@article{xu2025webbenchllmcodebenchmark,
  title={{Web}-{Bench}: A {LLM} Code Benchmark Based on Web Standards and Frameworks},
  author={Xu, K. and Mao, Y. and Guan, X. and Feng, Z.},
  journal={arXiv preprint arXiv:2505.07473},
  year={2025}
}

@article{adams2022combiningstateoftheartmodelsmaximal,
  title={Combining State-of-the-Art Models with Maximal Marginal Relevance for Few-Shot and Zero-Shot Multi-Document Summarization},
  author={Adams, D. and Suri, G. and Chali, Y.},
  journal={arXiv preprint arXiv:2211.10808},
  year={2022}
}



\end{document}